\documentclass{article}
\usepackage{spconf,amsmath,graphicx}
\usepackage{hyperref}
\usepackage{cleveref}
\usepackage{xpatch,xcolor}
\usepackage{makecell,booktabs}
\usepackage{amsmath,amssymb}
\usepackage{mathtools}
\usepackage{multicol,multirow}
\usepackage{array}
\usepackage{tabularx}
\usepackage{caption}
\usepackage{subcaption}
\usepackage{enumitem,kantlipsum}
\usepackage{cite}
\usepackage{url}

\hypersetup{
    colorlinks=true,
    linkcolor=purple,
    filecolor=magenta,      
    urlcolor=purple,
}
\newcolumntype{Y}{>{\centering\arraybackslash}X}
\newcolumntype{C}{ >{\centering\arraybackslash} m{4cm} }

\newcommand{\newpara}[1]{\vspace{1pt}\noindent\textbf{#1}}

\title{AugSumm: towards generalizable speech summarization using synthetic labels from large language models}

\name{Jee-weon Jung$^*$$^1$, Roshan Sharma\sthanks{$^*$Equal contribution.}$^1$, William Chen$^1$, Bhiksha Raj$^{1,2}$, Shinji Watanabe$^1$\thanks{Experiments of this work used the Bridges2 system at PSC and Delta system at NCSA through allocations CIS210014 and IRI120008P from the Advanced Cyberinfrastructure Coordination Ecosystem: Services \& Support (ACCESS) program, supported by National Science Foundation grants \#2138259,\#2138286, \#2138307, \#2137603, and \#2138296. Part of the ChatGPT augmentation was supported by KonanAI via OpenAI’s Research Access Program.}}
\address{Carnegie Mellon University, USA$^1$, Mohamed bin Zayed University of AI, Abu Dhabi$^2$}

\begin{document}
\ninept
\maketitle

\begin{abstract}
Abstractive speech summarization (SSUM) aims to generate human-like summaries from speech. 
Given variations in information captured and phrasing, recordings can be summarized in multiple ways. Therefore, it is more reasonable to consider a probabilistic distribution of all potential summaries rather than a single summary.
However, conventional SSUM models are mostly trained and evaluated with a single ground-truth (GT) human-annotated deterministic summary for every recording.
Generating multiple human references would be ideal to better represent the distribution statistically, but is impractical because annotation is expensive.  
We tackle this challenge by proposing \emph{AugSumm}, a method to leverage large language models (LLMs) as a proxy for human annotators to generate augmented summaries for training and evaluation.
First, we explore prompting strategies to generate synthetic summaries from ChatGPT. 
We validate the quality of synthetic summaries using multiple metrics including human evaluation, where we find that summaries generated using AugSumm are perceived as more valid to humans. 
Second, we develop methods to utilize synthetic summaries in training and evaluation. 
Experiments on How2 demonstrate that pre-training on synthetic summaries and fine-tuning on GT summaries improves ROUGE-L by 1 point on both GT and AugSumm-based test sets. 
AugSumm summaries are available at \texttt{\url{https://github.com/Jungjee/AugSumm}}.
\end{abstract}


\section{Introduction}
\label{sec:intro}
The goal of abstractive speech summarization (SSUM) ~\cite{murray2010interpretation,palaskar2019multimodal,li2019,shang2018unsupervised} is to condense essential information from long recordings and generate human-like summaries. Humans summarize speech in two steps: (i) select important semantic concepts and (ii) combine these concepts to form an abstractive summary. 
Humans differ in what they perceive as important~\cite{rath1961formation}, and how they combine these important concepts to form a summary. 
Therefore, numerous {\em valid} summaries exist given input speech due to differences in concept selection and semantic concept combination. 
These summaries can be reasonably modeled as belonging to a \textit{distribution}.
However, current models for SSUM use a deterministic single subjective ground-truth (GT) human-annotated summary from one annotator to train and evaluate SSUM models. We believe that sampling a single GT summary per sample does not sufficiently represent the entire distribution.
Hence both the training and evaluation of SSUM models with a single reference can be sub-optimal~\cite{cohan2022overview}.

Obtaining multiple human-annotated summaries for each recording would be ideal for a more accurate approximation of the underlying distribution. 
But, the high cost of human annotation makes it infeasible.  In this paper, we address this limitation by using large language models (LLMs) as a proxy for human annotators.
We use them as generative models to sample additional references for training and evaluation since they are exceptional at generating human-like text ~\cite{aher2023using,Sorensen_2022,argyle_busby_fulda_gubler_rytting_wingate_2023}.

To generate additional synthetic summaries (AugSumm)\footnote{\textbf{Aug}menting \textbf{Summ}aries.}, we leverage an existing SSUM corpus, How2\cite{sanabria2018how2} with GT transcript and summary, and the LLM ChatGPT (GPT3.5TURBO). 
We compare two methods : (a) \emph{direct summarization}, where we obtain the synthetic summaries based on the GT transcript, and (b) \emph{paraphrasing}, where we obtain synthetic summaries based on the GT summary. We further enforce the presence of semantic concepts in paraphrasing using ``concept words\footnote{A concept word here refers to an automatically extracted keyword.}'' within the prompt to guide factual and relevant summary generation from the LLM.

We also investigate approaches to leverage the augmented summaries in training and evaluation. 
For evaluation, we construct a new evaluation set with AugSumm summaries.
For training, we compare multiple approaches including (a) label data augmentation enlarging the number of samples in an epoch and (b) 2-stage training with pre-training and fine-tuning. 

We perform extensive experiments on the How2 dataset ~\cite{sanabria2018how2}. 
We first show that the LLM-generated AugSumm summaries are useful and valid by extensively evaluating the generated summaries on lexical-, semantic-, and human-labeling metrics. 
Our human evaluation finds that AugSumm summaries are more {\em valid} than GT summaries with a 95\% confidence interval.
We further demonstrate that AugSumm summaries can play a critical role in the trained SSUM model's behavior. Pre-training on both AugSumm and GT train sets, and fine-tuning on the GT set improved all 10 metrics across both GT and AugSumm test sets. 
In particular, the ROUGE-L~\cite{lin2004rouge} score was improved by more than 1 point absolute on both test sets.
The main contributions of this paper are:

\begin{enumerate}[leftmargin=*]
\item We propose AugSumm, a method to augment references for summarization using generative models for SSUM
\item We investigate two approaches to generate AugSumm summaries - direct and paraphrase summarization
\item We describe multiple approaches to use AugSumm summaries for training and evaluating SSUM models.
\end{enumerate}

\begin{table*}[t!]
  \caption{A comparison of the ground-truth human annotated summary and summaries generated with AugSumm methods. Paraphrase AugSumm leverages the ground-truth summary while direct AugSumm only uses the transcript.}
  \vspace{-8pt}
  \label{tab:prompt_ex}
  \begin{tabularx}{\linewidth}{l|X}
    \Xhline{1pt}
    Ground truth & The side angle squat pose is a great yoga exercise for pregnant \textcolor{red}{\textbf{women}} to increase leg strength and circulation as well as staying active and healthy throughout their third \textcolor{red}{\textbf{trimester}}. Stay a healthy pregnant woman with yoga poses and exercise tips from an experienced yoga instructor in this free video.\\
    \hline
    Concept words & side, angle, squat, pose, yoga, exercise, \textcolor{red}{\textbf{trimester}}, \textcolor{red}{\textbf{women}}, instructor, video\\
    \hline
    Paraphrased, w/ concept & The side angle squat pose is a beneficial yoga exercise for a pregnant \textcolor{red}{\textbf{woman}} in their first \textcolor{red}{\textbf{trimester}}. Stay active and healthy during pregnancy with yoga poses and exercise tips from an experienced yoga instructor in this free video.\\
    \hline
    Direct summarization & Katie and the instructor demonstrate a squat with side angle variation for pregnant \textcolor{red}{\textbf{women}} in their third \textcolor{red}{\textbf{trimester}}. They use a chair for support and point out that the added weight of the growing baby can make the exercise more challenging. They also demonstrate arm movements that engage the thigh muscles. The belly should be soft and breathing should be continuous. The pose can be modified to standing without the chair.\\
    \Xhline{1pt}
  \end{tabularx}
  \vspace{-10pt}
\end{table*}
\section{Methodology}
\subsection{End to End Speech Summarization}
\label{sec:e2e_ssum}
Traditionally, SSUM models are realized within a cascade framework by first transcribing long recordings using automatic speech recognition (ASR) and then summarizing transcripts using text summarization models.  Recently, it has been shown that end-to-end (E2E) models~\cite{sharma2022end,sharma2022xnor} outperform cascade models~\cite{matsuura2023transfer, matsuuraLeveraging,shon-etal-2023-slue}, particularly when ASR errors are present or useful para-linguistic information~\cite{kanoSpeech} is present.  E2E models are realized as attention-based encoder-decoder models and trained in two stages. In the first stage, the model is pre-trained to perform ASR. The intuition behind this is that ASR can produce semantically grounded speech representations that can be consequently refined for SSUM. Then, in the second stage, the trained ASR model is fine-tuned for SSUM. The model is optimized to maximize the log-likelihood of the reference summary using categorical cross-entropy.
Let $X$ be a sequence of speech features. An E2E SSUM model with parameters $\theta$ attempts to model $P(Y|X;\theta)$, by ingesting speech features $X$ to produce a sequence of summary tokens $Y$.

\begin{figure}[t]
\centering
    
\noindent\fbox{
\parbox{0.95\linewidth }{
    \footnotesize 
    \textit{You are here to create an extractive summary from the transcript. An extractive summary uses words from the input to convey the important portions of the video. Please make sure that the summary has between 40 and 60 words. Respond with only the extractive summary for: $\{$transcription$\}$.}
    \qquad \textit{\textbf{transcription}:} 
    
}}
    \begin{subfigure}[b]{\linewidth}
        \centering
        \vspace{-5pt}
        \caption{Direct Augsumm.}    
        \label{figsub:ds_word}
    \end{subfigure}
\fbox{
    \parbox{0.95\linewidth }{
    \footnotesize 
    \textit{You are here to paraphrase a given summary in the same style as the provided input. Please make sure that the summary has between {min(GT$\_$len-10, 20)} to {max(GT$\_$len+5, 20)} words.}
    \qquad \textit{\textbf{given summary}:}
}}
    \begin{subfigure}[b]{\linewidth}
        \centering
        \vspace{-5pt}
        \caption{Paraphrase AugSumm without concept words.}    
        \label{figsub:para_wo}
    \end{subfigure}

\fbox{%
    \parbox{0.95\linewidth }{%
    \footnotesize 
    \textit{You are here to paraphrase a given summary in the same style as the provided input. Please make sure that the summary has between {min(GT$\_$len-10, 20)} to {max(GT$\_$len+5, 20)} words. Also please include these words in the summary: $\{$important$\_$keys$\}$.}
    \qquad \textit{\textbf{given summary}:}
}}
    \begin{subfigure}[b]{\linewidth}
        \centering
        \vspace{-5pt}
        \caption{Paraphrase AugSumm with concept words. }    
        \label{figsub:para_w}
    \end{subfigure}

\vspace{-8pt}
\caption{Prompts for AugSumm.
Direct AugSumm produces word-level extractive summaries and is directly generated using only the transcript.
Paraphrase AugSumm requires GT summaries. {\em GT$\_$len} refers to the number of words of the GT summary.}

\label{fig:prompt}
\vspace{-10pt}
\end{figure}

\vspace{-7pt}
\subsection{Proposed: AugSumm}
\label{sec:augsumm}
\newpara{Formulation.} SSUM models attempt to represent the probability distribution $P(Y|X;\theta)$, i.e., the likelihood of the summary sequence $Y$ given the input speech $X$. To train SSUM models in a supervised manner, we sample pairs of $(X,Y)$ from the true joint distribution $\mathbb{P}(X,Y)$ of X and Y. The SSUM training dataset $D$ comprises such $(X,Y)$ pairs. 
To better approximate the true distribution $\mathbb{P}(X,Y)$, one would need to sample a large number of instances from it. Sampling in this case corresponds to obtaining more human annotations $Y$ for different speech inputs $X$. There are two main limitations of this sampling method. 
First, SSUM models are trained to estimate $P(Y|X;\theta)$. Random sampling from the joint distribution yields a set of valid $(X,Y)$ pairs, however, sampling from $P(Y|X)$ is likely better. The latter means that $X$ is selected first, and for every $X$, we sample multiple values from $P(Y|X)$, effectively improving distribution coverage and reflecting the fact that there are multiple valid summaries for every input. Second, additional samples require human annotations, and this is challenging to obtain in practice. 
The proposed AugSumm aims to sample additional training examples for every $X$ from the conditional distribution $\mathbb{P}(Y|X)$.  

Since the true probability density $\mathbb{P}(Y|X)$ is unknown, we resort to using a model with parameters $\theta$ that represents $P(Y|X;\theta)$ to obtain such samples. It is also challenging to obtain models that represent $P(Y|X)$, so we factor the conditional distribution to include the transcript $T$:  
\begin{equation}
\vspace{-6pt}
    P(Y|X) \approx \text{max}_T P(Y|T) P(T|X).
    \label{eqn:transcript}
\end{equation}

From Equation \ref{eqn:transcript}, the conditional distribution $P(Y|X)$ can be factored into the product of the likelihood of producing a summary given the transcript $P(Y|T)$, and the probability of obtaining the transcript $T$ from the input speech $X$. The second term $P(T|X)$ is 1 for GT transcripts (from ASR data or human transcription). Therefore, the problem of modeling $P(Y|X)$ reduces to modeling $P(Y|T)$. 
Any generative text summarization model, including an LLM, represents precisely this probability. 

Based on this formulation, we define AugSumm as obtaining synthetic summaries from the LLM by sampling as shown in Equation \ref{eqn:sampling}, where $P(Y | T, Prompt)$ represent the LLM operation with a static Prompt and $Y_{synth}$ represents the generated summary: 
\begin{equation}
\vspace{-4pt}
    Y_{synth} \sim P(Y|T, Prompt).
    \label{eqn:sampling}
    \vspace{-2pt}
\end{equation}

\newpara{Direct AugSumm.}
\label{ssec:summarize}
The first approach, {\em direct} AugSumm, utilizes Equation \ref{eqn:sampling} to generate synthetic summaries $Y_{synth}$ from transcripts by using $direct\_prompt$ in place of $Prompt$:
\begin{equation}
\vspace{-4pt}
    Y_{synth} \sim P(Y|T, direct\_prompt).
    \vspace{-2pt}
\end{equation}
However, because the proposed AugSumm adopts LLMs in place of humans to obtain summaries, the output summary can be erroneous. To address this, we prompt the LLM to generate extractive summaries\footnote{Extractive summaries are obtained by selecting key phrases and sentences from the input text, and then combining them.} using words that already exist in the transcript rather than generating unconstrained abstractive summaries. 
Figure \ref{figsub:ds_word} displays the prompt used.
Direct AugSumm can be used with any speech data with available transcriptions and hence it has the potential to train SSUM models with existing thousands of hours of ASR corpora. 

\newpara{Paraphrase AugSumm.}
\label{ssec:paraphrase}
Direct AugSumm presents a straightforward approach to sample additional summaries. 
However, we empirically found that current LLMs do not stably generate valid abstractive summaries.
As reported in prior work, prompting affects the generated outputs of LLM significantly~\cite{white2023prompt}. 
We therefore propose another method, {\em paraphrase} AugSumm to sample additional summaries.
In paraphrase AugSumm, we devise prompts to query LLMs so that they paraphrase GT reference summaries. An example prompt is shown in Figure ~\ref{figsub:para_wo}.

Mathematically put, paraphrase AugSumm represents the conditional probability $P(Y|T)$ based on the conditional density of $P(Y,Y_{GT}|T)$ as shown in Equation \ref{eqn:paraphrase}: 
\begin{equation}
\vspace{-4pt}
  \begin{aligned}
     P(Y|T) = \sum_{Y_{GT}} P(Y,Y_{GT}|T) & = \sum_{Y_{GT}} P(Y|T,Y_{GT})P(Y_{GT}|T)\\
     & \approx \sum_{Y_{GT}} P(Y|Y_{GT})P(Y_{GT}|T).
     \label{eqn:paraphrase}
  \end{aligned}
\vspace{2pt}
\end{equation}
Here, $P(Y_{GT}|T)$ represents the process of human annotation and is a constant.
Using the fact that $Y$ and $Y_{GT}$ are conditionally independent given the transcript $T$, we note that $P(Y|T)$ can be expressed in terms of $P(Y|Y_{GT})$. This probability can be modeled by an LLM that produces a $Y$ based only on the GT reference summary. 
Here, using a different static prompt $paraphrase\_prompt$, the LLM models: 
\begin{equation}
\vspace{-4pt}
    Y_{synth} \sim P(Y|Y_{GT}, paraphrase\_prompt),
\end{equation}
for paraphrase generation. Compared to direct AugSumm, paraphrase AugSumm uses $Y_{GT}$ instead of $T$ and a different prompt.

On top of paraphrase AugSumm, we further propose to use a specific set of semantic concepts. 
For this purpose, we first extract semantic concepts as noun/verb phrases from the summary following ~\cite{palaskar2021multimodal,sharma2022end}. 
Then, we modify the prompt as shown in Figure ~\ref{figsub:para_w} to produce the extracted semantic concepts within the paraphrase AugSumm. 

\begin{table}[t!]
  \caption{Quality comparison of paraphrase AugSumm. Results between 2 to 7 columns are model-based where the first four are from UniEval~\cite{zhong2022towards} and the two last are from BERTscore~\cite{zhang2019bertscore}. ROUGE-1 is based on lexical similarity. (Pa: paraphrase, Coh: coherence, Con: consistency, Flu: fluency, Rel: relevance, RG1: ROUGE-1 with transcript, BT-T: BERTscore with transcript, BT-G: BERTscore with ground-truth summary, HME: human evaluation (validity, \%)}
  \vspace{-5pt}
  \label{tab:quality_paraphrase}
  \scalebox{0.85}{
  \begin{tabular}{c|cccccc|c|c}
    \Xhline{1pt}
     & \textbf{Coh} & \textbf{Con} & \textbf{Flu} & \textbf{Rel} & \textbf{BT-T} & \textbf{BT-G} & \textbf{RG1} & \textbf{HME}\\
    \Xhline{1pt}
    GT & 81.11 & 80.10 & 91.30 & - & 82.65 & - & 12.48 & 51.67\\
    Pa & 86.04 & 83.76 & 94.89 & 98.32 & 82.65 & 93.77 & 12.49 &65.33 \\
    \Xhline{1pt}
  \end{tabular}}
  \vspace{-10pt}
\end{table}
\subsection{Training and Evaluating with Synthetic Summaries}

\label{ssec:training_tech}
We explore multiple approaches to utilize synthetic AugSumm summaries. A simple baseline \emph{Augsumm only}, trains the SSUM model on pairs comprising real speech and augmented summaries. Another approach adopts synthetic summaries as additional reference labels in training by {\em enlarging} the training set. Such an approach would combine the impacts of training with synthetic references and GT references. 
If the goal of summarization is to optimize proximity with either synthetic labels only or GT labels only, but not both, then, a {\em 2-stage training} paradigm can be adopted. In this approach, the SSUM model is first pre-trained on the non-target labels (synthetic if the objective is to maximize proximity with GT labels and vice-versa) and then fine-tuned with the target labels (the labels we wish to optimize proximity with). 
We also sample $Y_{synth}$ from the test set to construct an AugSumm test set and report models' performance on this set.

\vspace{-12pt}
\section{Experimental setup}
\vspace{-5pt}
\label{sec:exp}

\begin{table}[t!]
  \caption{Quality comparison of direct AugSumm. (Di: direct AugSumm, RGL: ROUGE-L with transcript)}
  \vspace{-8pt}
  \label{tab:quality_direct}
  \begin{tabularx}{\linewidth}{c|YYYY|YY}
    \Xhline{1pt}
     & \textbf{Coh} & \textbf{Con} & \textbf{Flu} & \textbf{BT-T} & \textbf{RG1} & \textbf{RGL}\\
    \Xhline{1pt}
    GT & 81.11 & 80.10 & 91.30 & 82.65 & 12.48 & 8.30\\
    Di & 97.00 & 93.04 & 92.57 & 85.09  & 28.65 & 23.15\\
    \Xhline{1pt}
  \end{tabularx}
  \vspace{-10pt}
\end{table}
\begin{table*}[t!]
  \caption{Baseline results trained with and without AugSumm labels. Results better than the Baseline are depicted in boldface and the best performance is underlined. ROUGE metrics with $*$ are measured using AugSumm labels. (BT-A: BERTscore with AugSumm test set)}
  \vspace{-8pt}
  \label{tab:base_and_augonly}
  \begin{tabularx}{\textwidth}{l|YYYYY|YY|YYY}
    \Xhline{1pt}
    & \textbf{Coh} & \textbf{Con} & \textbf{Flu} & \textbf{Rel} &  \textbf{BT-G} & \textbf{RG1} & \textbf{RGL} &  \textbf{RG1*} & \textbf{RGL*} & \textbf{BT-A}\\
    \Xhline{1pt}
    Sharma et al.,~\cite{sharma2022end} & - & - & - & - & 91.53 & 60.73 & 56.10 & - & - & -\\
    Baseline (ours, w/o AugSumm) & 68.58 & 72.66 & 82.92 & 77.57 & \underline{\textbf{92.16}} & \underline{\textbf{61.95}} & \underline{\textbf{57.52}} & \underline{\textbf{47.69}} & \underline{\textbf{36.77}} & \underline{\textbf{88.40}}\\
    \hline
    Paraphrase, w/o concept & \textbf{77.18} & \textbf{81.09} & \underline{\textbf{86.81}} & \textbf{81.93} & 88.88 & 51.44 & 35.20 & 43.98 & 30.13 & 87.16\\
    Paraphrase, w/ concept & \textbf{78.26} & \textbf{81.29} & \textbf{85.54} & \underline{\textbf{82.65}} & 90.21 & 54.95 & 35.69 & 47.80 & 31.78 & \underline{\textbf{88.40}}\\
    Direct & \textbf{81.73} & \textbf{86.37} & \textbf{83.15} & 62.93 & 81.36 & 23.04 & 16.10 & 17.99 & 12.84 & 80.69\\
    \Xhline{1pt}
  \end{tabularx}
  \vspace{-10pt}
\end{table*}
\begin{table*}[t!]
  \caption{Comparing approaches to utilize AugSumm for training. All experiments have been conducted using ``Paraphrase, w/ concept'' in Table~\ref{tab:base_and_augonly}. Results better than the Baseline is depicted in boldface and the best performance is underlined. (Pt: pre-train, Ft: fine-tune)}
  \vspace{-8pt}
  \label{tab:augsumm_approaches}
  \begin{tabularx}{\textwidth}{l|YYYYY|YY|YYY}
    \Xhline{1pt}
    & \textbf{Coh} & \textbf{Con} & \textbf{Flu} & \textbf{Rel} &  \textbf{BT-G} & \textbf{RG1} & \textbf{RGL} & \textbf{RG1*} & \textbf{RGL*} & \textbf{BT-A} \\
    \Xhline{1pt}
    & \multicolumn{8}{c}{1 stage training}\\
    \hline\hline
    GT-only (=Baseline) & 68.58 & 72.66 & 82.92 & 77.57 & 92.16 & 61.95 & 57.52 & 47.69 & 36.77 & 88.40\\
    AugSumm-only & \underline{\textbf{78.26}} & \underline{\textbf{81.29}} & \underline{\textbf{85.54}} & \underline{\textbf{82.65}} & 90.21 & 54.95 & 35.69 & \textbf{47.80} & 31.78 & 88.40\\
    GT half + AugSumm half & \textbf{73.34} & \textbf{76.77} & \textbf{83.91} & \textbf{80.10} &91.32 & 58.89 & 47.76 & 47.94 & 34.89 & \textbf{88.49}\\
    Enlarge (GT+AugSumm) & \textbf{75.73} & \textbf{78.38} & \textbf{84.81} & \textbf{82.18} & 91.52 & 60.30 & 47.30 & \textbf{\underline{49.14}} & 34.54 & \textbf{88.66}\\
    \Xhline{1pt}
    & \multicolumn{8}{c}{2 stage training}\\
    \hline\hline
    Pt Augsumm-only $\to$ Ft GT & \textbf{69.74} & \textbf{73.52} & \textbf{83.48} & \textbf{77.11} & \textbf{92.44} & \textbf{63.17} & \textbf{58.96} & \textbf{48.50} & \textbf{\underline{37.65}} & \textbf{\underline{88.59}}\\
    Pt Enlarge $\to$ Ft GT & \textbf{69.70} & \textbf{73.42} & \textbf{83.17} & \textbf{78.79} & \underline{\textbf{92.48}} & \underline{\textbf{63.64}} & \underline{\textbf{59.37}} & \textbf{48.52} & \textbf{37.61} & \textbf{88.53}\\
    Pt GT $\to$ Ft AugSumm & 67.18 & 72.33 & 82.01 & 76.18 & \textbf{92.20} & \textbf{63.00} & \textbf{58.76} & \textbf{47.74} & \textbf{36.79} & 88.27\\
    \hline
    
    \Xhline{1pt}
  \end{tabularx}
  \vspace{-6pt}
\end{table*}

\subsection{Dataset: How2-2000h}
\label{ssec:db}
The How2 dataset includes approximately 2000 hours of instructional videos crawled from YouTube~\cite{sanabria2018how2}. 
It has English transcriptions and user-provided descriptions that describe how to perform particular tasks, like repairing a bicycle or cooking a particular dish. We utilize the official training, validation, and test splits.

\subsection{Model and training configurations}
\label{ssec:model_cfg}

Our E2E SSUM models are implemented as attention-based sequence models with a 12-layer conformer \cite{gulati20_conformer,guo2021recent} encoder, and a 6-layer transformer decoder. The encoder and decoder use a feedforward dimension of 2,048 and 512, along with 8 and 4 attention heads respectively. To efficiently process longer sequences, we replace standard self-attention with a parameter-less Fourier transform from FNet \cite{lee2022fnet}, which has been shown to work well for SSUM \cite{kanoSpeech}. 43-dimensional filter-bank pitch features are extracted using Kaldi ~\cite{kaldi} and used as input to the model. 
The model is first pre-trained on ASR using the hybrid CTC/attention loss \cite{watanabeHybrid} with a CTC weight of 0.3 before being trained for SSUM using cross-entropy. 
SpecAugment ~\cite{park19e_interspeech} is used for ASR pre-training, while no augmentation is used for SSUM training.

\newpara{Reproducibility.}
To make our work reproducible given the stochasticity of LLMs, we make public the source code\footnote{\url{https://github.com/espnet/espnet}.} and AugSumm summary data.\footnote{\url{https://github.com/Jungjee/AugSumm}.}

\vspace{-10pt}
\subsection{Metrics}
\label{ssec:metrics}
We employ diverse sets of metrics which can be divided into lexical and semantic. 
Lexical metrics in this paper include ROUGE-1~\cite{lin2004rouge} and ROUGE-L~\cite{lin2004rouge}. 
These metrics measure in a word sense how close the model-generated summaries are compared to the reference summary.
However, like most metrics, they are sub-optimal in some cases. 
Semantic metrics complement lexical metrics by directly comparing semantic embeddings to assess similarity.
In this work, we employ BERTscore~\cite{zhang2019bertscore} and UniEval~\cite{zhong2022towards} as model-based metrics. 
We selected these two model-based metrics because BERTscore is a widely adopted metric in the research community and UniEval, although relatively new, provides scores across various dimensions that are also used in human evaluation -- coherence, consistency, fluency, and relevance.
Coherence and consistency compare the model-generated output with the transcript. Fluency is measured using the model-generated output only. Relevance compares the model-generated output with a reference.

\section{Experimental Results}
\label{sec:results}
\vspace{-7pt}
\newpara{Quality of AugSumm summaries.}
Tables \ref{tab:quality_paraphrase} and \ref{tab:quality_direct} report multiple metrics that reflect the quality of LLM-generated summaries.
For paraphrase AugSumm, \Cref{tab:quality_paraphrase}, all three UniEval metrics except relevance were higher than the GT. For relevance, which measures how relevant a given summary is compared to the GT, is 98.32. Surprisingly, even BERTscores computed with ASR transcript as reference were almost identical, indicating that semantics are preserved. ROUGE-1 scores are also similar, showing a similar degree of word overlap in synthetic summaries. 
For direct AugSumm, \Cref{tab:quality_direct}, three UniEval metrics were higher than both the paraphrase AugSumm and GT.\footnote{As summaries being extractive, it's not meaningful to measure relevance.}
ROUGE-1 and -L are higher than GT. High ROUGE values occur due to the fact that we constrain the model to use the words within the transcript to generate direct summaries. 
We further conducted a {\em human evaluation}, with 20 annotators and 15 questions each to choose among the four options: (i) summary 1 is valid while summary 2 is not, (ii)  summary 2 is valid while summary 1 is not, (iii) Both summaries are valid, and (iv) Both summaries are invalid. In total, we received 300 responses, with 51, 92, 104, and 53 responses for each respective option. Here, summary 1 is the GT How2 summary and 2 is the AugSumm summary. Therefore, from human ratings,  GT and AugSumm summaries were considered valid  51.67 $\pm$ 0.0565\% and 65.33  $\pm$ 0.0538\%  of the time respectively with a 95\% confidence interval. We partly attribute this to the fact that How2 summaries were constructed by crawling user descriptions from YouTube and inserting them into predefined sentence structures. In summary, AugSumm summaries are valid based on lexical, semantic, and human evaluation.

\newpara{Baselines and AugSumm-only.}
Table~\ref{tab:base_and_augonly} compares the baseline and SSUM models trained only with AugSumm summaries. Our baseline based on FNet outperforms the reported performance with restricted self-attention \cite{sharma2022end},  and thus, we use the former.
All models trained with AugSumm demonstrated improved performance on UniEval; however, the baseline has the best result on other metrics. 
Overall, we conclude that models trained with AugSumm have the potential for improvement when jointly trained with an existing GT summary using various techniques.

\newpara{Techniques utilizing AugSumm.}
Table~\ref{tab:augsumm_approaches} shows the result of our various techniques to leverage AugSumm along with the existing GT summary.
``GT half + AugSumm half'' and ``Enlarge'' in Rows 3 and 4 show results when using both AugSumm and GT labels by mixing the two or using them all.
These two experiments improve all four UniEval metrics, but ROUGE or BERTscore with the GT summary does not improve.
We contend that this is because, in this setup, models are optimized to produce both the GT or AugSumm summary during training, both of which contain different semantic concepts and phrase structures. ROUGE and BERTScore on the AugSumm test set improve when using both GT and AugSumm data in training. 

Rows 5 to 7 demonstrate the impact of adopting 2-stage training. 
2-stage training is effective regardless of what kind of label is used in the pre-train and fine-tuning stage. 
Among them, pre-training with both labels and fine-tuning with the GT summary (Row 6) performed the best. All 10 metrics showed improvement and ROUGE scores improved on both GT and AugSumm test sets. 
Based on these results, we conclude that AugSumm can be used to derive SSUM models that generalize better. 

\vspace{-5pt}
\section{Conclusion and future directions}
\label{sec:conclusion}
In this paper, we proposed \emph{AugSumm}, which uses LLMs to augment summary labels for SSUM. 
We compared two methods to generate synthetic reference summaries - directly summarizing speech transcripts and paraphrasing existing GT summaries. Experiments on How2 showed that both were beneficial. 
Multiple methods to utilize augmented and GT summaries were discussed, with the 2-stage training framework yielding improvements across lexical and semantic metrics and across original and augmented test sets.
We can obtain a 1-point absolute improvement in ROUGE-L by using AugSumm.

The success of this work, using LLMs as a source of data augmentation, inspires multiple directions for future work. 
First, advanced training techniques such as contrastive learning and mix-up can be now applied to better leverage multiple summary labels per audio file.  Second, this method can be extended to several tasks such as speech translation where the label is a sentence.

\vspace{-9pt}

\clearpage
\bibliographystyle{IEEEbib}
\bibliography{strings,refs}

\begin{thebibliography}{10}

\bibitem{murray2010interpretation}
Gabriel Murray, Giuseppe Carenini, and Raymond Ng,
\newblock ``Interpretation and transformation for abstracting conversations,''
\newblock in {\em Proc. NAACL}, 2010, pp. 894--902.

\bibitem{palaskar2019multimodal}
Shruti Palaskar, Jind{\v{r}}ich Libovick{\'y}, Spandana Gella, and Florian
  Metze,
\newblock ``Multimodal abstractive summarization for how2 videos,''
\newblock in {\em Proc. ACL}, July 2019.

\bibitem{li2019}
Haoran Li, Junnan Zhu, Cong Ma, Jiajun Zhang, and Chengqing Zong,
\newblock ``Read, watch, listen, and summarize: Multi-modal summarization for
  asynchronous text, image, audio and video,''
\newblock {\em IEEE Transactions on Knowledge and Data Engineering}, vol. 31,
  no. 5, pp. 996--1009, 2019.

\bibitem{shang2018unsupervised}
Guokan Shang, Wensi Ding, Zekun Zhang, Antoine Jean-Pierre Tixier, Polykarpos
  Meladianos, Michalis Vazirgiannis, and Jean-Pierre Lorr{\'e},
\newblock ``Unsupervised abstractive meeting summarization with multi-sentence
  compression and budgeted submodular maximization,''
\newblock in {\em Proc. NAACL}, 2018.

\bibitem{rath1961formation}
GJ~Rath, A~Resnick, and Terry~R Savage,
\newblock ``The formation of abstracts by the selection of sentences. part i.
  sentence selection by men and machines,''
\newblock {\em American Documentation}, vol. 12, no. 2, pp. 139--141, 1961.

\bibitem{cohan2022overview}
Arman Cohan, Guy Feigenblat, Tirthankar Ghosal, and Michal Shmueli-Scheuer,
\newblock ``Overview of the first shared task on multi perspective scientific
  document summarization (mup),''
\newblock in {\em Proceedings of the Third Workshop on Scholarly Document
  Processing}, 2022, pp. 263--267.

\bibitem{aher2023using}
Gati~V Aher, Rosa~I Arriaga, and Adam~Tauman Kalai,
\newblock ``Using large language models to simulate multiple humans and
  replicate human subject studies,''
\newblock in {\em Proc. ICML}, 2023.

\bibitem{Sorensen_2022}
Taylor Sorensen, Joshua Robinson, Christopher Rytting, Alexander Shaw, Kyle
  Rogers, Alexia Delorey, Mahmoud Khalil, Nancy Fulda, and David Wingate,
\newblock ``An information-theoretic approach to prompt engineering without
  ground truth labels,''
\newblock in {\em Proc. ACL}. 2022, Association for Computational Linguistics.

\bibitem{argyle_busby_fulda_gubler_rytting_wingate_2023}
Lisa~P. Argyle, Ethan~C. Busby, Nancy Fulda, Joshua~R. Gubler, Christopher
  Rytting, and David Wingate,
\newblock ``Out of one, many: Using language models to simulate human
  samples,''
\newblock {\em Political Analysis}, vol. 31, no. 3, pp. 337–351, 2023.

\bibitem{sanabria2018how2}
Ramon Sanabria, Ozan Caglayan, Shruti Palaskar, Desmond Elliott, Lo{\"\i}c
  Barrault, Lucia Specia, and Florian Metze,
\newblock ``How2: A large-scale dataset for multimodal language
  understanding,''
\newblock in {\em Proc. NeurIPS}, 2018.

\bibitem{lin2004rouge}
Chin-Yew Lin,
\newblock ``Rouge: A package for automatic evaluation of summaries,''
\newblock in {\em Proc. ACL}, 2004.

\bibitem{sharma2022end}
Roshan Sharma, Shruti Palaskar, Alan~W Black, and Florian Metze,
\newblock ``End-to-end speech summarization using restricted self-attention,''
\newblock in {\em Proc. ICASSP}, 2022.

\bibitem{sharma2022xnor}
Roshan Sharma and Bhiksha Raj,
\newblock ``Xnor-former: Learning accurate approximations in long speech
  transformers,''
\newblock {\em arXiv preprint arXiv:2210.16643}, 2022.

\bibitem{matsuura2023transfer}
Kohei Matsuura, Takanori Ashihara, Takafumi Moriya, Tomohiro Tanaka, Takatomo
  Kano, Atsunori Ogawa, and Marc Delcroix,
\newblock ``Transfer learning from pre-trained language models improves
  end-to-end speech summarization,''
\newblock in {\em Proc. Interspeech}, 2023.

\bibitem{matsuuraLeveraging}
Kohei Matsuura, Takanori Ashihara, Takafumi Moriya, Tomohiro Tanaka, Atsunori
  Ogawa, Marc Delcroix, and Ryo Masumura,
\newblock ``Leveraging large text corpora for end-to-end speech
  summarization,''
\newblock in {\em Proc. ICASSP}, 2023.

\bibitem{shon-etal-2023-slue}
Suwon Shon, Siddhant Arora, Chyi-Jiunn Lin, Ankita Pasad, Felix Wu, Roshan~S
  Sharma, Wei-Lun Wu, Hung-yi Lee, Karen Livescu, and Shinji Watanabe,
\newblock ``{SLUE} phase-2: A benchmark suite of diverse spoken language
  understanding tasks,''
\newblock in {\em Proc. ACL}, 2023.

\bibitem{kanoSpeech}
Takatomo Kano, Atsunori Ogawa, Marc Delcroix, Roshan Sharma, Kohei Matsuura,
  and Shinji Watanabe,
\newblock ``Speech summarization of long spoken document: Improving memory
  efficiency of speech/text encoders,''
\newblock in {\em Proc. ICASSP}, 2023.

\bibitem{white2023prompt}
Jules White, Quchen Fu, Sam Hays, Michael Sandborn, Carlos Olea, Henry Gilbert,
  Ashraf Elnashar, Jesse Spencer-Smith, and Douglas~C Schmidt,
\newblock ``A prompt pattern catalog to enhance prompt engineering with
  chatgpt,''
\newblock {\em arXiv preprint arXiv:2302.11382}, 2023.

\bibitem{palaskar2021multimodal}
Shruti Palaskar, Ruslan Salakhutdinov, Alan~W Black, and Florian Metze,
\newblock ``Multimodal speech summarization through semantic concept
  learning.,''
\newblock in {\em Proc. Interspeech}, 2021, pp. 791--795.

\bibitem{zhong2022towards}
Ming Zhong, Yang Liu, Da~Yin, Yuning Mao, Yizhu Jiao, Pengfei Liu, Chenguang
  Zhu, Heng Ji, and Jiawei Han,
\newblock ``Towards a unified multi-dimensional evaluator for text
  generation,''
\newblock in {\em Proc. EMNLP}, 2022.

\bibitem{zhang2019bertscore}
Tianyi Zhang, Varsha Kishore, Felix Wu, Kilian~Q Weinberger, and Yoav Artzi,
\newblock ``Bertscore: Evaluating text generation with bert,''
\newblock in {\em Proc. ICLR}, 2019.

\bibitem{gulati20_conformer}
Anmol Gulati, James Qin, Chung-Cheng Chiu, Niki Parmar, Yu~Zhang, Jiahui Yu,
  Wei Han, Shibo Wang, Zhengdong Zhang, Yonghui Wu, and Ruoming Pang,
\newblock ``{Conformer: Convolution-augmented Transformer for Speech
  Recognition},''
\newblock in {\em Proc. Interspeech 2020}, 2020.

\bibitem{guo2021recent}
Pengcheng Guo, Florian Boyer, Xuankai Chang, Tomoki Hayashi, Yosuke Higuchi,
  Hirofumi Inaguma, Naoyuki Kamo, Chenda Li, Daniel Garcia-Romero, Jiatong Shi,
  et~al.,
\newblock ``Recent developments on espnet toolkit boosted by conformer,''
\newblock in {\em Proc. ICASSP}, 2021.

\bibitem{lee2022fnet}
James Lee-Thorp, Joshua Ainslie, Ilya Eckstein, and Santiago Ontanon,
\newblock ``Fnet: Mixing tokens with fourier transforms,''
\newblock in {\em Proc. NAACL}, 2022.

\bibitem{kaldi}
Daniel Povey, Arnab Ghoshal, Gilles Boulianne, Lukas Burget, Ondrej Glembek,
  Nagendra Goel, Mirko Hannemann, Petr Motlicek, Yanmin Qian, Petr Schwarz, Jan
  Silovsky, Georg Stemmer, and Karel Vesely,
\newblock ``The kaldi speech recognition toolkit,''
\newblock in {\em Proc. ASRU}, Dec. 2011.

\bibitem{watanabeHybrid}
Shinji Watanabe, Takaaki Hori, Suyoun Kim, John~R. Hershey, and Tomoki Hayashi,
\newblock ``Hybrid {CTC}/attention architecture for end-to-end speech
  recognition,''
\newblock {\em IEEE Journal of Selected Topics in Signal Processing}, vol. 11,
  no. 8, pp. 1240--1253, 2017.

\bibitem{park19e_interspeech}
Daniel~S. Park, William Chan, Yu~Zhang, Chung-Cheng Chiu, Barret Zoph, Ekin~D.
  Cubuk, and Quoc~V. Le,
\newblock ``{SpecAugment: A Simple Data Augmentation Method for Automatic
  Speech Recognition},''
\newblock in {\em Proc. Interspeech}, 2019, pp. 2613--2617.

\end{thebibliography}

\end{document}